\documentclass[10pt,twocolumn,letterpaper]{article}

\usepackage{wacv}
\usepackage{times}
\usepackage{epsfig}
\usepackage{graphicx}
\usepackage{amsmath}
\usepackage{amssymb}

\usepackage{url}            
\usepackage{hhline}
\usepackage{multirow}
\usepackage{amsfonts}       
\usepackage{color}

\usepackage[pagebackref=true,breaklinks=true,letterpaper=true,colorlinks,bookmarks=false]{hyperref}

\newcommand{\todo}[1]{{}}

\wacvfinalcopy 

\def\wacvPaperID{236} 

\ifwacvfinal\fi
\begin{document}

\title{Improved Mixed-Example Data Augmentation}

\author{
  Cecilia Summers \\
  Department of Computer Science\\
  University of Auckland\\
  {\tt\small cecilia.summers.07@gmail.com} \\
\and
  Michael J. Dinneen \\
  Department of Computer Science\\
  University of Auckland\\
  {\tt\small mjd@cs.auckland.ac.nz} \\
}

\maketitle
\ifwacvfinal\thispagestyle{empty}\fi

\begin{abstract}
In order to reduce overfitting, neural networks are typically trained with data augmentation, the practice of artificially generating additional training data via label-preserving transformations of existing training examples.
While these types of transformations make intuitive sense, recent work has demonstrated that even non-label-preserving data augmentation can be surprisingly effective, examining this type of data augmentation through linear combinations of pairs of examples.
Despite their effectiveness, little is known about why such methods work.
In this work, we aim to explore a new, more generalized form of this type of data augmentation in order to determine whether such linearity is necessary. 
By considering this broader scope of ``mixed-example data augmentation'', we find a much larger space of practical augmentation techniques, including methods that improve upon previous state-of-the-art.
This generalization has benefits beyond the promise of improved performance, revealing a number of types of mixed-example data augmentation that are radically different from those considered in prior work, which provides evidence that current theories for the effectiveness of such methods are incomplete and suggests that any such theory must explain a much broader phenomenon. Code is available at \url{https://github.com/ceciliaresearch/MixedExample}.
\end{abstract}

\vspace{-4mm}
\section{Introduction}

\begin{figure}[t]
  \includegraphics[width=.98\linewidth]{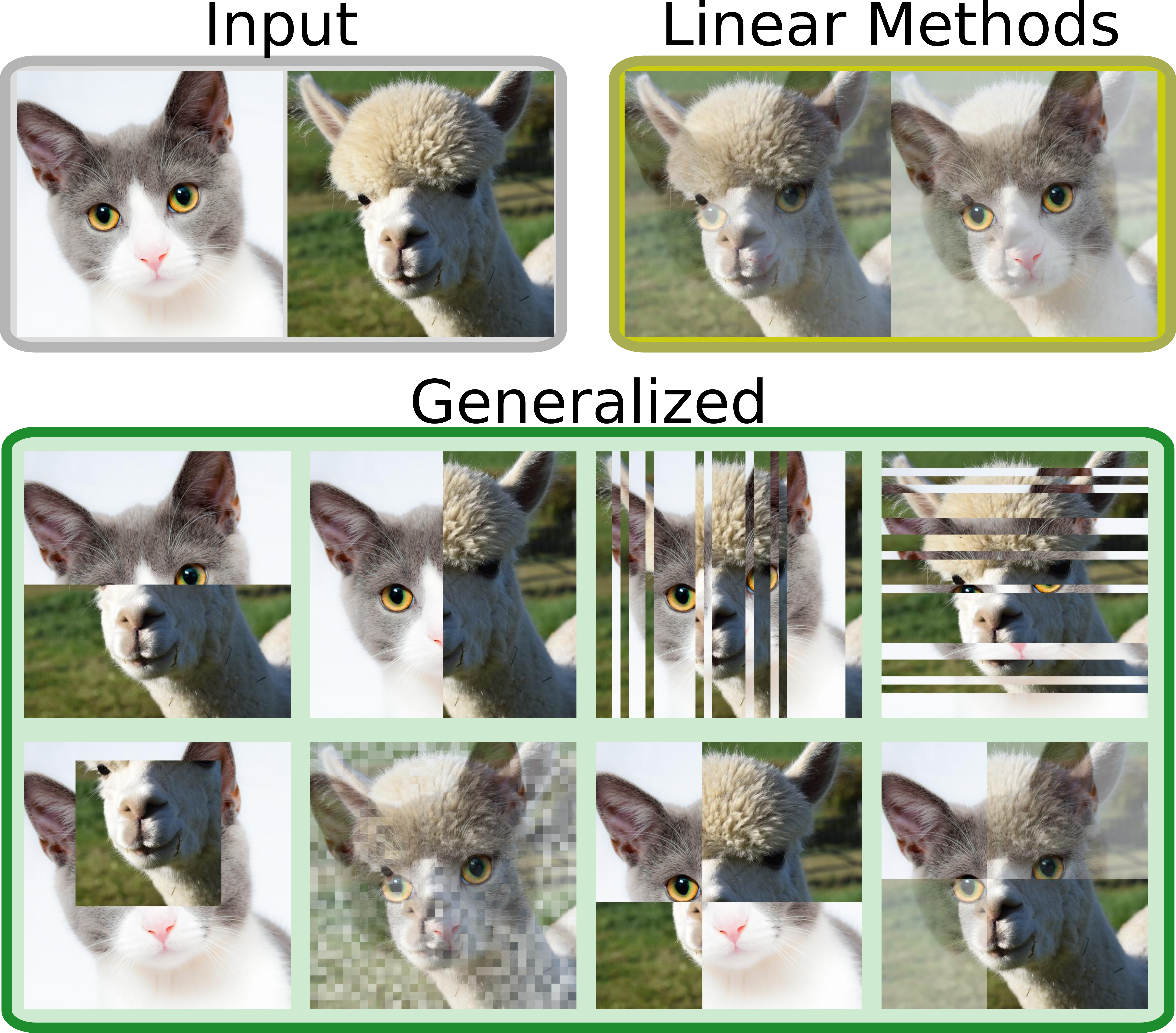}
  \caption{
Given two input examples, mixed-example data augmentation consists of the set of functions that combine the inputs into a novel image, perhaps with an appearance unrealistic to humans. Previous work has considered the case of linear transformations, \emph{i.e.} element-wise weighted averaging. In this work, we explore a more generalized space of functions in order to determine whether non-linear functions are similarly effective. Shown above are examples of 8 of the methods considered in this work, all of which result in improved performance over non-mixed example augmentation, illustrating that the space of viable functions is much broader than previously realized.
}
  \label{fig:fig1}
  \vspace{-3mm}
\end{figure}


Deep neural networks have demonstrated remarkable performance on many tasks previously considered intractable, but they come with a cost: they require large amounts of data. While great progress has been made at making neural nets more data-efficient through the use of improved network architectures and training methods, the limitation remains, with the greatest effect in data-starved domains such as robotics~\cite{levine2016end,munawar2017spatio} and medical applications~\cite{esteva2017dermatologist,gulshan2016development,gu2017melanoma}.

To get around the requirement for large amounts of data, machine learning practitioners typically either transfer knowledge from other tasks, or employ large amounts of ``data augmentation'', the practice of generating synthetic data based on real examples. This synthetic data artificially expands the size of datasets used for training models by many orders of magnitude, and typically takes the form of transformations that maintain the realistic appearance of the input, \emph{e.g.} for image classification, flipping an image horizontally or making minor adjustments to its brightness.

A handful of recent work, however, has pointed out a new direction: data augmentation can take the form of mixing training examples together, specifically via element-wise averages of pairs of inputs~\cite{zhang2017mixup,tokozume2018image,tokozume2018audio,inoue2018data}. Though this type of data augmentation does not produce data realistic to humans, it is surprisingly effective at training models, significantly improving performance across a variety of tasks and domains even \emph{after} other forms of data augmentation are considered.
Due to the nascency of such techniques, we currently do not have a good understanding of why they are effective. One hypothesis comes from Zhang et al.~\cite{zhang2017mixup}, who suggest that the added linearity this form of data augmentation encourages is a useful inductive bias. Another is from Tokozume et al.~\cite{tokozume2018image}, who put forth the idea that CNNs treat imagery as waveform data and that such data augmentation puts constraints on internal feature distributions. While the true modus operandi by which these methods are successful remains an area of active research, these works nevertheless point to new directions for reducing the dependence of modern deep neural networks on large quantities of labeled data.

In this paper, we ask the question: is linearity critical to the success of these data augmentation methods that combine multiple input examples?
To answer this question, we explore this new area of data augmentation as generalized functions of multiple inputs, which we denote ``mixed-example data augmentation''. We propose a variety of alternative methods for example generation, and surprisingly find that almost all methods result in improvements over models trained without any form of mixed-example data augmentation (see Fig.~\ref{fig:fig1} for examples). As a byproduct of increasing this search space, we find multiple methods that improve upon existing work~\cite{zhang2017mixup,tokozume2018image}. Though we do not propose new theoretical reasons for the utility of mixed-example data augmentation, our experiments shed empirical light on existing hypotheses, providing evidence that they are incomplete.

The remainder of this paper is organized as follows: In Sec.~\ref{sec:related_work} we review related work in data augmentation and regularization more generally, and in Sec.~\ref{sec:methods} we present our exploration of mixed-example data augmentation, overviewing 14 alternatives. We present experiments on CIFAR-10, CIFAR-100, and Caltech-256 in Sec.~\ref{sec:experiments}, then conclude with a discussion in Sec.~\ref{sec:discussion}.

\section{Related Work}
\label{sec:related_work}
\subparagraph{Data Augmentation.}
Data augmentation is key to the success of most modern neural networks across nearly every domain.
Here we limit our discussion of data augmentation to applications with images, the focus of this work.

Common forms of data augmentation include random crops, horizontal flipping, and color augmentation~\cite{krizhevsky2012imagenet}, which improve robustness to translation, reflection, and illumination, respectively. Occasionally, random scalings are also done~\cite{simonyan2014very} as well as random rotations and affine transformations, though these tend to get somewhat less use. One more recent form of data augmentation consists of zeroing out random parts of the image~\cite{devries2017improved,zhong2017random}, a structured form of input-layer dropout, which works surprisingly well on the CIFAR-10/100 datasets.
Though they differ greatly in technique, all of these methods are a form of label-preserving data augmentation, \emph{i.e.} they are designed to maintain the label of the transformed image, which requires a small amount of task-specific knowledge. For example, in the context of image classification, an image of a cat, even after horizontally flipping it and altering its brightness, is still recognizable as an image of a cat.

More directly relevant to our work is recent progress on generating images for training as linear combinations of other training images~\cite{zhang2017mixup,tokozume2018image,inoue2018data}. While we save a detailed overview for Sec.~\ref{sec:linearity}, these methods take the general form of randomly generating a mixing coefficient and producing a new image and label as a convex combination of two images/labels. Tokozume et al.~\cite{tokozume2018image} further considered an improved form based on the interpretation of images as waveform data, and Inoue restricted the sampling coefficient to $0.5$.
Our research, inspired by these, considers more generalized functions of a different form, showing that the space of effective mixed-example data augmentation is much more broad than linearity-based methods.

\vspace{-4mm}

\subparagraph{Regularization.}
Closely related to the topic of data augmentation is regularization. One of the most common approaches to regularization is weight decay~\cite{krogh1992simple}, which is equivalent to $L_2$-regularization when using a vanilla stochastic gradient descent learning rule. Other common types of regularization include Dropout~\cite{srivastava2014dropout}, which can also be considered a form of feature-space data augmentation injected at intermediate layers in a neural network, and Batch Normalization~\cite{ioffe2015batch}, which has a regularization effect due to randomness in minibatch statistics. Other more exotic forms of regularization include randomly dropping out layers~\cite{huang2016deep} and introducing disparities between forward- and backward-propagation~\cite{gastaldi2017shake}.
Data augmentation and regularization can both be viewed as ways to incorporate prior knowledge into models, either via invariances in data (label-preserving data augmentation) or through priors on how weights and activations in neural networks should behave (regularization), and both have the goal of reducing the train-test generalization gap.
As such, it is folk wisdom that there is a tradeoff between the optimal amount of data augmentation and regularization to use --- for example, Zhang et. al~\cite{zhang2017mixup} found that using \emph{mixup} well required a 5x lower amount of weight decay than without \emph{mixup} on CIFAR-10. 

\section{Methods}
\label{sec:methods}

\subsection{General Formulation}
Most uses of label-preserving data augmentation can be represented by stochastic functions of the form
\begin{equation}
  \label{eq:label-preserving}
  (\tilde{x}, \tilde{y}) = \tilde{f}(x,y) = (f(x), y) .
\end{equation}
For example, $f$ may be a function randomly altering the brightness of its input, horizontally flipping it, or applying a projective transformation. Of key note, however, is that $\tilde{f}$ is an identity function with respect to its second input.

In this work, we consider generalized methods for data augmentation of the form:

\begin{equation}
  \label{eq:two_examples}
  (\tilde{x}, \tilde{y}) = \tilde{f}(\{(x_i, y_i)\}_{i=1}^2)
\end{equation}

That is, we consider arbitrary functions mapping two examples into a single new training example. This is a strict generalization of the form of augmentation in Eq.~\ref{eq:label-preserving}, which can be obtained by ignoring either one of the two inputs. In theory one could consider functions with $N > 2$ examples as input, but in initial experiments (also agreeing with \cite{zhang2017mixup}) we did not see improvement beyond $N=2$, so in this work we restrict our methods to this setting.

\subsection{Linearity-Based Methods}
\label{sec:linearity}
In this section we present previous work~\cite{zhang2017mixup,tokozume2018image,tokozume2018audio,inoue2018data}, which can be represented as special cases of Eq.~\ref{eq:two_examples} in which $\tilde{f}$ is a linear combination of $(x_1, y_1), (x_2, y_2)$:

\subparagraph{Mixup.} In \emph{mixup}\cite{zhang2017mixup}, the augmentation function $\tilde{f}$ is represented by:

\begin{equation}
\begin{split}
\tilde{x} &	= \lambda x_1 + (1 - \lambda) x_2 \\
\tilde{y}	& = \lambda y_1 + (1 - \lambda) y_2
\end{split}
\end{equation}

where $\lambda \sim Beta(\alpha, \alpha)$ for each pair of examples, with $\alpha$ a hyperparameter. For experiments on CIFAR-10 and CIFAR-100, Zhang et al.~\cite{zhang2017mixup} used the value $\alpha = 1$, which results in a uniform distribution between 0 and 1, and found that on larger datasets such as ImageNet~\cite{russakovsky2015imagenet} a smaller value of $\alpha$ was required due to underfitting.  \emph{Mixup} was motivated as encouraging linearity between training examples, with the hypothesis that linearity is an effective inductive bias (an assumption built into a model) for most models. Indeed, \emph{mixup} was shown to be useful across a wide variety of tasks and models. As we shall show in our work, this picture is incomplete --- linearity, while useful, is not required for mixed-example data augmentation to be effective, and extremely non-linear methods can perform nearly as well.

\subparagraph{Between Class (BC+).}
Tokozume \emph{et al.}~\cite{tokozume2018image} developed two methods for mixed-example data augmentation. The first, ``BC'' (Between-Class), is equivalent to \emph{mixup} and was developed in parallel with Zhang \emph{et al.}~\cite{zhang2017mixup}. Improving upon ``BC'' and building upon their previous work with audio~\cite{tokozume2018audio}, Tokozume \emph{et al.} developed ``BC+'', which is based on the intuition that neural networks (specifically CNNs) can treat imagery as waveform data. Using this intuition, two improvements were made:

First, waveforms are naturally zero-mean signals, while images are not. Therefore, Tokozume \emph{et al.} first subtract each image's mean (computed across all channels, \emph{i.e.} a single number per image) from itself before further processing. This stands in contrast to most recent work with CNNs~\cite{xie2017aggregated}, which uses a mean across an entire dataset for normalization.

The second improvement comes from noting that combining examples in a strictly linear fashion does not produce a \emph{perceptually} linear combination of images, a fact which was an acute concern in their prior work on audio~\cite{tokozume2018audio}. To solve this, Tokozume \emph{et al.} use the standard deviation of each image $\sigma_1, \sigma_2$ to measure energy, and ultimately derive the mixing equation

\begin{equation}
\begin{split}
  \tilde{x} = \frac{p (x_1 - \mu_1) + (1 -p) (x_2 - \mu_2)}{\sqrt{p^2 + (1 - p)^2}} \\
  \text{where}\quad p = \frac{1}{1 + \frac{\sigma_1}{\sigma_2} \cdot \frac{1 - \lambda}{\lambda}}
\end{split}
\end{equation}

with $\lambda \sim U[0,1]$ and the label being linearly determined as in \emph{mixup}. While BC+ is technically a non-linear method, we group it together with linearity-based methods such as \emph{mixup} since the non-linearity only occurs in the normalization term and the method is still fundamentally based on element-wise averaging.

\subsection{Non-Linear Methods}
\label{sec:nonlinear}
\begin{figure}[t]
  \includegraphics[width=.98\linewidth]{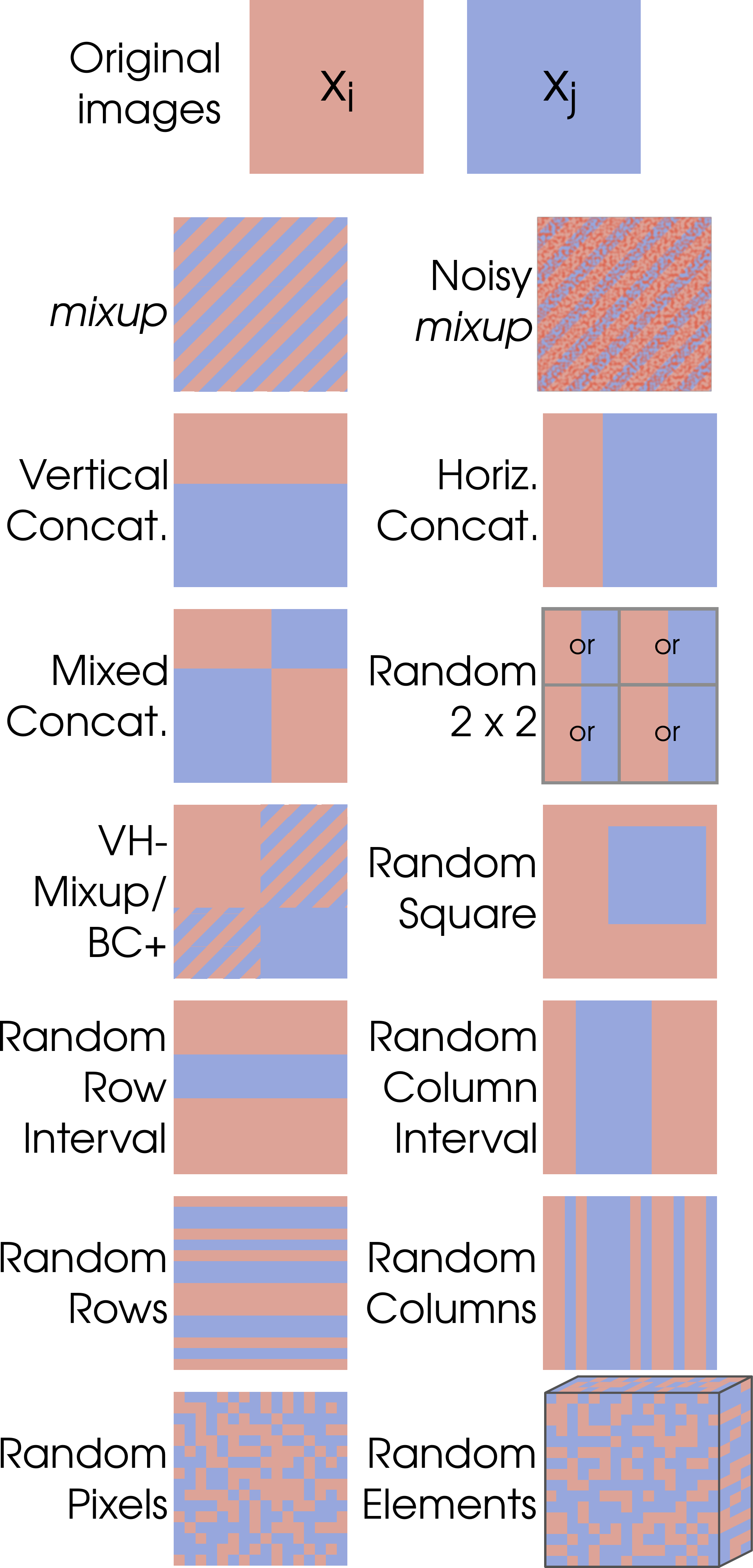}
  \caption{Example outputs for each mixed-example data augmentation method. See text (Sec.~\ref{sec:methods}) for details of each method. Diagonal stripes are used to indicate element-wise weighted averaging. Note that for ``VH-Mixup/VH-BC+'', the bottom-left and upper-right region use different weights for the weighted average depending on $\lambda_3$ (see text).
}
  \label{fig:methods}
\end{figure}

We now illustrate the generality of our formulation (Eq.~\ref{eq:two_examples}) by presenting many different non-linear methods for mixed-example data augmentation. Although we present them as alternatives to linearity-based approaches, we note that most of these methods are largely orthogonal both to such approaches as well as to more traditional forms of data augmentation and can potentially be employed in combination with them. Illustrative examples are shown in Fig.~\ref{fig:methods}.

\subparagraph{Vertical Concat.} As in \emph{mixup} and similar to BC+, ``Vertical Concat'' begins be sampling a random mixing coefficient $\lambda \sim Beta(\alpha, \alpha)$. However, instead of element-wise averaging, in this method the top $\lambda$ fraction of image $x_1$ is vertically concatenated with the bottom $(1 - \lambda)$ fraction of image $x_2$. Formally, we have

\begin{equation}
\tilde{x}(r,c) = \left\{\begin{array}{lr}
x_1(r,c), & \text{if } r \leq \lambda H \\
x_2(r,c) & \text{otherwise}
\end{array} \right.
\end{equation}

where $H$ is the height of the image and $x(r,c)$ denotes the 3-dimensional pixel at row $r$ and column $c$ of an image $x$.
Though simple, this is an extremely non-linear transformation with respect to the input. The label $\tilde{y}$ remains equal to the original labels weighted by the mixing coefficient: $\lambda y_1 + (1 - \lambda) y_2$. Thus, a network trained with ``Vertical Concat'' must not only correctly classify the top and bottom portions of the image, but also correctly identify what fraction of the image they occupy.

\subparagraph{Horizontal Concat.} This method is similar to ``Vertical Concat.'', but instead horizontally concatenates the left $\lambda$ fraction of $x_1$ with the right $(1 - \lambda)$ fraction of $x_2$:

\begin{equation}
\tilde{x}(r,c) = \left\{\begin{array}{lr}
x_1(r,c), & \text{if } c \leq \lambda W \\
x_2(r,c) & \text{otherwise}
\end{array} \right.
\end{equation}

where $W$ is the width of the image. As before, we have $\tilde{y} = \lambda y_1 + (1 - \lambda) y_2$.

\subparagraph{Mixed Concat.} This is a combination of vertical and horizontal concatenation: first we sample $\lambda_1, \lambda_2 \sim Beta(\alpha, \alpha)$. Then we divide the output image in a $2 \times 2$ grid as shown in Fig.~\ref{fig:methods}, where the horizontal boundary between grid members is determined by $\lambda_1$ and the vertical boundary is determined by $\lambda_2$. The top-left and bottom-right portions of the output image are set to the corresponding pixel values in $x_1$, and the top-right and bottom-left are set to $x_2$, with $\tilde{y} = (\lambda_1 \lambda_2 + (1 - \lambda_1)(1 - \lambda_2)) y_1 + (\lambda_1 (1 - \lambda_2) + (1 - \lambda_1)\lambda_2) y_2$, \emph{i.e.} $\tilde{y}$ is determined by the relative area of $x_1$ vs $x_2$. Another interpretation of ``Mixed Concat.'' is an application of ``Vertical Concat.'' to two images produced by ``Horizontal Concat.'' with the same mixing coefficient but opposite argument order.

\subparagraph{Random $\mathbf{2 \times 2}$.} This method is a more randomized version of ``Mixed Concat.''. First, this method divides the image into a $2 \times 2$ grid with random sizes as before, but instead of using a fixed assignment of grid cells to input images, ``Random $2 \times 2$'' randomly decides for each square in the grid whether it should take content from $x_1$ or $x_2$. This prevents the network from relying on the fixed assignment in ``Mixed Concat.``, instead forcing it to adapt to potentially changing positions of the input content.  The target label $\tilde{y}$ is measured as a function of the relative area of $x_1$ vs $x_2$ in the generated image.

One implementation detail that we have found to modestly help ``Random $2 \times 2$'' is a constraint on the $2 \times 2$ grid produced. Specifically, this constraint forces the intersection lines of the grid to occur in the middle $p$ fraction of the image, preventing image content from becoming too long, narrow, or not even present. Though this constraint is not critical to the success of the method, we find that it tends to improve performance by a small but significant amount. In our experiments we set $p$ to $0.5$.

\subparagraph{VH-Mixup.}
In order to explore whether it is possible to combine the strengths of non-linear methods with methods based on linearity, we introduce ``VH-Mixup'', the goal of which is to leverage the advantages of ``Vertical Concat.'', ``Horizontal Concat.'', and \emph{mixup}~\cite{zhang2017mixup} (or equivalently BC~\cite{tokozume2018image}).
First, two intermediate images are made as the result of ``Vertical Concat.'' and ``Horizontal Concat.'', each with their own randomly chosen $\lambda$. Then, \emph{mixup} is applied with these two images as input. This has the effect of producing an image where the top-left is from $x_1$, the bottom-right is from $x_2$, and the top-right and bottom-left are mixed between the two with different mixing coefficients. All together, with $\lambda_1, \lambda_2, \lambda_3 \sim Beta(\alpha, \alpha)$, we have that $\tilde{x}(r,c)$ is equal to:

\begin{equation}
\begin{array}{lr}
x_1(r,c), & \text{if } r \leq \lambda_1 H \wedge c \leq \lambda_2 W \\
  \lambda_3 x_1(r,c) + (1 - \lambda_3) x_2(r,c), & \text{if } r \leq \lambda_1 H \wedge c > \lambda_2 W \\
  (1 - \lambda_3) x_1(r,c) + \lambda_3 x_2(r,c), & \text{if } r > \lambda_1 H \wedge c \leq \lambda_2 W \\
x_2(r,c) & \text{if } r > \lambda_1 H \wedge c > \lambda_2 W
\end{array}
\end{equation}

The label $\tilde{y}$ is determined in a straightforward manner based on the rules for label generation in ``Vertical Concat.'', ``Horizontal Concat.'', and \emph{mixup}, and can be thought of as the expected fraction of its value a random pixel takes from $x_1$ vs $x_2$.

\subparagraph{VH-BC+.} Rather than combining the outputs of ``Vertical Concat.'' and ``Horizontal Concat.'' with \emph{mixup}, in ``VH-BC+'' they are combined with BC+. While it is tempting to think of this as simply replacing \emph{mixup} with BC+, there is a subtle implementation detail: when to subtract the mean for each image. There are two options: directly before applying BC+, \emph{i.e.} after performing ``Vertical Concat.'' and ``Horizontal Concat.'', or before producing either of the concatenated images. We argue that the latter is correct --- if an output of one of the concatenation methods is made into a zero-mean image, then it will have relatively little effect, as the image will still be clearly made of two distinct parts, even only based on first-order statistics (the mean). However, if the original two images are zero-meaned before either of the concatenation methods, then each concatenated image will still be zero mean in expectation, but the boundary between the two will no longer be as easily discerned.  Indeed, we tested both methods in initial experiments, and while we found both to perform reasonably, ultimately the latter was slightly better on average.

\subparagraph{Random Square.} In this method, a random square within $x_1$ is replaced with a portion of $x_2$. This method is inspired by Cutout~\cite{devries2017improved} and random erasing data augmentation~\cite{zhong2017random}, but instead of replacing the subimage with 0, we replace it with part of a different image. As in Cutout, the size of the square is a hyperparameter, which we set to 16 pixels. One subtle implementation choice with this method is which region in $x_2$ to use as a replacement --- in our implementation, we use a portion of $x_2$ picked randomly among all regions of the appropriate size, which we found slightly better than using the region in $x_2$ directly corresponding to the replaced region of $x_1$.

\subparagraph{Random Column Interval.} This method is a slight generalization of ``Horizontal Concat.'' --- a random interval of columns is picked and that part of image $x_1$ is replaced with columns in $x_2$. The difference between this method and ``Horizontal Concat.'' is that this column interval need not begin with the first column. The interval in this method is picked by sampling the lower bound of the interval uniformly, with the upper bound then sampled uniformly between all possible remaining upper bounds.

\subparagraph{Random Row Interval.} This method is identical to ``Random Column Interval'' but is applied to a random interval of rows instead of columns.

\subparagraph{Random Rows.} For each row in the output image $\tilde{x}$, this method randomly samples whether to take the row from $x_1$ or $x_2$, where the probability of choosing the corresponding row in $x_1$ is given by $\lambda$. As before, $\tilde{y}$ is determined based on the fraction of rows that were taken from $x_1$ compared with $x_2$. One interpretation of this method is as a higher-frequency variant of ``Vertical Concat.'' in that rows are still taken either entirely from $x_1$ or $x_2$, but with this method they may alternate between $x_1$ and $x_2$, possibly many times, rather than being grouped into a single large block of rows.

\subparagraph{Random Columns.} This method is identical to ``Random Rows'' but samples columns instead of rows.

\subparagraph{Random Pixels.} This method is similar to ``Random Rows'' but samples each pixel separately. That is, after first sampling $\lambda$, a matrix of size $W \times H$ is created, consisting of numbers drawn uniformly from $[0,1]$, and converted to a binary matrix via comparison with $\lambda$. This is interpreted as a boolean mask which can be element-wise multipled with $x_1$ and $x_2$ in order to efficiently compute the output. The label $\tilde{y}$ can also be easily determined using the expected value of the mask.

\subparagraph{Random Elements.} This method is similar to ``Random Rows'' but samples each element of the image separately. That is, when the image is represented as a $H \times W \times 3$ tensor (using RGB), each element in the tensor is randomly sampled from the corresponding value in either $x_1$ or $x_2$. As with ``Random Pixels'', this can be efficiently computed by using a $H \times W \times 3$ tensor of random numbers.

\subparagraph{Noisy Mixup.}
Normally, in \emph{mixup}~\cite{zhang2017mixup}, a single $\lambda \sim Beta(\alpha, \alpha)$ is sampled and then used across the entire image (or $\lambda \sim U[0,1]$ for BC~\cite{tokozume2018image}). However, in order to produce an output with an expected label of $\lambda y_1 + (1 - \lambda) y_2$, there is no need for $\lambda$ to be the same across an entire image -- instead, as long as its expectation is the same, the same label applies. In this method, we first sample $\lambda$ in the same fashion, but then for each pixel identified by a row $r$ and column $c$ we add random zero-centered noise to the mixing coefficient: $\lambda_{r,c} = \lambda + \ell_{r,c}$ where $\ell_{r,c} \sim N(0, \sigma^2)$, with $\sigma^2$ a hyperparameter that we set to $\sqrt{0.025} \approx 0.16$, indicating adding a small amount of pixel-wise data-dependent noise. It is also useful to constrain $\lambda_{r,c}$ to lie in the range $[0,1]$, \emph{i.e.} $\lambda_{r,c} = \max{(\min{(\lambda + \ell_{r,c}, 1)}, 0)}$. Though this method is nearly linear, we find that it makes for an interesting comparison experimentally with the strictly linear methods \emph{mixup}~\cite{zhang2017mixup} and BC~\cite{tokozume2018image}.

\section{Experiments}
\label{sec:experiments}
\subsection{Implementation Details}

In all experiments, we perform mixed-example data augmentation by directly pairing together two examples at a time, rather than doing it on a batch-by-batch basis~\cite{zhang2017mixup}. While this has the potential to slow down data processing due to twice as much I/O and non-vectorized operations, it is somewhat simpler to develop with, especially for the slightly more involved data generation methods. Furthermore, we found that we were still able to generate data fast enough for models to use via effective use of dataset caching, which keeps I/O to a minimum. Initial experiments furthermore suggested that this does not affect accuracy. We also found that it was crucial to do other types of data augmentation (\emph{e.g.} random cropping and flips) before applying any type of mixed-example data augmentation, with differences in accuracy greater than $1\%$ on CIFAR-10, but do not currently have an explanation for why this is important, an implementation detail we also found shared in all existing open-source code of prior work. All experiments were done on a desktop with two Nvidia Geforce GTX 1080 Ti GPUs. We now list dataset-specific implementation details.

\subparagraph{CIFAR 10/100.} We perform the bulk of our experiments on the CIFAR-10 and CIFAR-100 datasets~\cite{krizhevsky2009learning}, the primary test bed used by prior work~\cite{zhang2017mixup,tokozume2018image}.  Following \cite{zhang2017mixup}, we conduct our experiments using the pre-activation ResNet-18~\cite{he2016identity}, which we re-implemented in TensorFlow~\cite{abadi2016tensorflow}. This network architecture has the advantage of having a relatively high accuracy (\emph{e.g.} $5.4\%$ error on CIFAR-10) while taking only 2 hours for a complete training run with any of the methods. Furthermore, previous work~\cite{zhang2017mixup,tokozume2018image} has already shown strong correlations in improvements across model architectures. We also note that this ResNet variant is slightly different from the official pre-activation ResNet-18, attaining somewhat higher accuracy.

Following both Zhang \emph{et al.}~\cite{zhang2017mixup} and Tokozume \emph{et al.}~\cite{tokozume2018image}, we use $\alpha=1$ where applicable, which results in a uniform distribution for $\lambda$, though this parameter can in principle be tuned based on the extent of overfitting. On CIFAR-10 we use a weight decay of $10^{-4}$ for all mixed-example methods and $5 \cdot 10^{-4}$ for the baseline ResNet-18, and on CIFAR-100 we use a weight decay of $5 \cdot 10^{-4}$ for all methods on CIFAR-100, which we found necessary in order to reproduce prior work~\cite{zhang2017mixup}, making a difference of slightly more than $1\%$ in final accuracy.

Following \cite{he2016identity}, we use minibatches of size 128, the learning rate starts at .01 for a warm-up period of 400 steps, increases to 0.1, then decays by a factor of 10 after 32,000, 48,000, and 70,000 steps. In practice, we noticed that this learning rate strategy can be somewhat unstable, with losses spiking up at the 400-step transition, after which models failed to recover well and ended up with a few percent lower accuracy than they would otherwise. Though the extent of this problem depended on the method, for reproducibility we have taken the practice of running 20 copies of each model for three epochs ($\approx$ 1,200 steps) and then only continuing the three models with the lowest loss values, which tended to not observe dramatic spikes in loss. This both improved final performance and reduced training variability.

\subparagraph{Caltech-256.} 
For experiments on Caltech-256~\cite{griffin2007caltech} we used the Inception-v3~\cite{szegedy2016rethinking} architecture with the default ``Inception'' preprocessing, resulting in a $299 \times 299$ pixel image input to the model. We used the default weight decay for Inception models of  $4 \cdot 10^{-5}$ and a batch size of 64 for all methods. For the baseline model we used a learning rate of .03, decayed by a factor of 10 when validation performance saturated, which occurred after 20,000 and 26,000 steps. The learning rate additionally had a warm-up period of 2,000 steps, during which time we increased it at a log-linear rate from $3 \cdot 10^{-5}$ to .03. All other methods had a similar warm-up phase and initial learning rate, but with learning rate decays by a factor of 10 after 45,000 and 57,000 steps, echoing the observation in prior work~\cite{zhang2017mixup,tokozume2018image} that mixed-example data augmentation can take longer to train, particularly with larger models. For ``BC+'' and ``VH-BC+'', we found it important to add a small constant when determining the standard deviation of each image in order to avoid numerical issues that made the loss diverge.

\subsection{Results}

\begin{table}
  \begin{centering}
  \begin{tabular}{|c|r|c|}
		\hline
		\multirow{ 18}{*}{CIFAR-10} &
    Method & Error (\%) \\ \cline{2-3} \hhline{|~:==|}
    & ResNet-18 & 5.4 \\ \cline{2-3}
    & \emph{mixup}\cite{zhang2017mixup} & 4.3 \\  \cline{2-3}
    & BC+\cite{tokozume2018image} & 4.2 \\  \cline{2-3} \hhline{|~:==|}
    & Rand. Elems. & 6.2 \\ \cline{2-3}
    & Rand. Pixels & 5.7 \\ \cline{2-3}
    & Rand. Col. Int. & 5.1 \\ \cline{2-3}
    & Rand. Cols & 4.8 \\ \cline{2-3}
    & Horiz. Concat. & 4.7 \\ \cline{2-3}
    & Rand. Rows & 4.6 \\ \cline{2-3}
    & Noisy Mixup & 4.5 \\ \cline{2-3}
    & Rand. Row. Int. & 4.5 \\ \cline{2-3}
    & Vert. Concat. & 4.4 \\ \cline{2-3}
    & Mixed. Concat. & 4.4 \\ \cline{2-3}
    & Rand. Square. & 4.3 \\ \cline{2-3}
    & \emph{Rand. $2 \times 2$} & \emph{4.1} \\ \cline{2-3}
    & \emph{VH-BC+} & \emph{3.8} \\ \cline{2-3}
    & \emph{VH-Mixup} & \emph{3.8} \\ \cline{2-3}
		\hline
  \end{tabular}
    \caption{Experimental results on CIFAR-10. All numbers are the average across three training runs, measured at the final step of training, and methods are ordered by performance. Numbers for the baseline ResNet-18 model, \emph{mixup}, and BC+ are from our TensorFlow re-implementation. Italicized method names and performances indicate methods which performed better than either existing state-of-the-art mixed-example method.}
  \label{table:cifar10}
  \end{centering}
\end{table}

\subparagraph{CIFAR-10.}
CIFAR-10~\cite{krizhevsky2009learning} consists of 60,000 images of size $32 \times 32$ pixels, split evenly among 10 categories, with 50,000 training images and 10,000 test images, and is a standard test bed for training of small-scale deep learning models, having been used extensively in related work~\cite{zhang2017mixup,tokozume2018image}. Results on CIFAR-10 are shown in Table~\ref{table:cifar10}. A few trends are immediately apparent:

First, we examine the central question of our work: is linearity required for mixed-example data augmentation to be successful? Our experimental results answer this clearly: \emph{linearity is not required for effective mixed-example data augmentation}. Rather, the space of useful mixed-example data augmentation appears to be much larger than realized in previous work~\cite{zhang2017mixup,tokozume2018image,inoue2018data} --- with the exception of ``Rand. Pixels'' and ``Rand. Elems'', all other mixed-example techniques improved upon the baseline ResNet. Even the simplest of methods, ``Horiz. Concat'' and ``Vert Concat', improved upon the baseline significantly, and are perhaps the least similar to prior work of the methods considered.

While linearity may not be a requirement in order for a method to improve upon baseline performance, is it required among the most effective methods? Again, our results indicate that this need not be the case. While ``VH-BC+'' and ``VH-Mixup'' have some element of linearity in portions of the image, ``Rand. $2 \times 2$'', despite containing no element-wise weighted averaging at all, was just as useful a form of data augmentation as \emph{BC+} and \emph{mixup}, even slightly outperforming them in the sample set of runs we conducted. While we agree with previous work that linearity on its own can be a fruitful inductive bias, it is by no means necessary.

Can we get the best of both worlds by combining the insights of linearity as an inductive bias with non-linear types of mixed-example data augmentation?
Two of the methods we explored, ``VH-Mixup'' and ``VH-BC+'', do just that, and in fact both were able to outperform all other approaches, setting a new state of the art for mixed-example data augmentation. This result is particularly promising due to the nascency of mixed-example approaches and the general applicability to a wide range of tasks (for the methods in this paper, tasks in computer vision). 

Last, in an effort to learn more about which aspects of such augmentation methods are useful, it is worth remarking on the methods that did not work well as negative examples.
In particular, ``Rand. Elems'' and ``Rand. Pixels'' both worked worse than the baseline of doing no mixed-example data augmentation. These methods have a commonality: by treating every pixel differently, they exhibit the tendency to introduce high-frequency signals in the data. We hypothesize that this type of data augmentation makes it more difficult for models to capture local details within images, forcing them to rely more on low-frequency content and limiting their ability to properly learn from all available signals. In a similar vein, we also note that ``Noisy Mixup'', although it worked reasonably well, was not even as effective as \emph{mixup} without any modifications, which we also attribute to the addition of high-frequency content.

\begin{table}
  \begin{centering}
  \begin{tabular}{|c|r|c|}
		\hline
\multirow{ 18}{*}{CIFAR-100} &
    Method & Error (\%) \\ \cline{2-3} \hhline{|~:==|}
    & ResNet-18 & 23.6 \\ \cline{2-3}
    & \emph{mixup}\cite{zhang2017mixup} & 21.3 \\ \cline{2-3}
    & BC+\cite{tokozume2018image} & 21.1 \\  \cline{2-3} \hhline{|~:==|}
    & Rand. Elems. & 24.2 \\ \cline{2-3}
    & Rand. Pixels & 24.0 \\ \cline{2-3}
    & Rand. Cols & 22.4 \\ \cline{2-3}
    & Noisy Mixup & 21.8 \\ \cline{2-3}
    & Horiz. Concat. & 21.7 \\ \cline{2-3}
    & Rand. Col. Int. & 21.4 \\ \cline{2-3}
    & \emph{Rand. Square.} & \emph{20.9} \\ \cline{2-3}
    & \emph{Rand. Rows} & \emph{20.9} \\ \cline{2-3}
    & \emph{Mixed. Concat.} & \emph{20.9} \\ \cline{2-3}
    & \emph{Vert. Concat.} & \emph{20.8} \\ \cline{2-3}
    & \emph{Rand. $2 \times 2$} & \emph{20.4} \\ \cline{2-3}
    & \emph{Rand. Row. Int.} & \emph{20.1} \\ \cline{2-3}
    & \emph{VH-BC+} & \emph{19.9} \\ \cline{2-3}
    & \emph{VH-Mixup} & \emph{19.7} \\ \cline{2-3}
		\hline
  \end{tabular}
    \caption{Experimental results on CIFAR-100. As in CIFAR-10, all numbers are the average across three training runs, measured at the final step of training, and methods are ordered by performance. Italics indicates better than existing state-of-the-art mixed-example methods.}
  \label{table:cifar100}
  \end{centering}
\end{table}

\subparagraph{CIFAR-100.}
CIFAR-100~\cite{krizhevsky2009learning} is a 100-class companion of CIFAR-10 with otherwise similar properties. We present our results on CIFAR-100 in Table~\ref{table:cifar100} Trends on CIFAR-100 were largely similar to results on CIFAR-10, with the best and worst methods consistent, though the ordering in between changed somewhat. One clear difference, though, is that many more of our exploratory methods outperformed prior work on CIFAR-100 (8 for CIFAR-100 vs 3 for CIFAR-10). While we do not offer any compelling hypothesis for this change, it provides evidence that at least some minor differences in effectiveness between each mixed-example method are likely to be data-dependent.

A further point worth noting, shared across both CIFAR-10 and CIFAR-100, is that row-based methods performed better than their column-based counterparts: ``Vert. Concat'' outperformed ``Horiz. Concat'',
``Rand. Rows'' outperformed ``Rand. Cols'', and ``Rand. Row. Int'' improved upon ``Rand. Col. Int''. This potentially indicates the importance of keeping horizontal information intact when doing data augmentation, a finding reminiscent of much older work in picking horizontally-shaped spatial pooling grids~\cite{lazebnik2006beyond}.

\subparagraph{Caltech-256.}
In order to test methods for mixed-example data augmentation on larger, more real-world images, we additionally evaluate on Caltech-256~\cite{griffin2007caltech}, a dataset of 256 categories. Since there is no canonical training, validation, or test splits, we constructed splits by randomly taking 40 images from each category for training, 10 for validation, and 30 for testing, resulting in splits of size 10,240, 2560, and 7,680, respectively. While smaller than other datasets of large natural images, \emph{e.g.} ImageNet~\cite{russakovsky2015imagenet}, experiments are also much more tractable, taking roughly 10 hours on average when training from scratch, compared with an estimated 25 days to run a single ImageNet experiment, which is prohibitively long.
For this set of experiments, we focused our analysis on the best-performing methods from CIFAR-10 and 100, ``VH-Mixup'' and ``VH-BC+'', in addition to the three baselines. Results are presented in Table~\ref{table:caltech256}.

\begin{table}
  \begin{centering}
  \begin{tabular}{|c|r|c|}
		\hline
\multirow{ 6}{*}{Caltech-256} &
    Method & Accuracy (\%) \\ \cline{2-3} \hhline{|~:==|}
    & Inception-v3 & 48.6 \\ \cline{2-3}
    & \emph{mixup}\cite{zhang2017mixup} & 57.3 \\ \cline{2-3}
    & BC+\cite{tokozume2018image} & 57.4 \\  \cline{2-3} \hhline{|~:==|}
    & VH-Mixup & 56.3 \\ \cline{2-3}
    & \emph{VH-BC+} & \emph{59.7} \\ \cline{2-3}
		\hline
  \end{tabular}
    \caption{Experimental results on Caltech-256. Accuracies are determined by a single run of model training, with evaluation checkpoints picked based on maximum validation performance.}
  \label{table:caltech256}
  \end{centering}
\end{table}

Most noticeably, we found that \emph{all} mixed-example data augmentation methods were able to improve performance over the baseline Inception-v3 network. The effect is dramatic, with improvements of up to $10\%$ in accuracy above baseline. This highlights the strength of mixed-example methods, particularly with high-capacity models, a finding that strengthens results from prior work~\cite{zhang2017mixup,tokozume2018image}. 

Within the set of mixed-example methods, though, ordering is less obvious --- while it is clear that ``VH-BC+'' was particularly successful, the reason by which it was so much better than ``VH-Mixup'' remains mysterious. It is also worth noting that confidence intervals for these experiments are somewhat wide: a $95\%$ confidence interval due to data sampling alone is roughly $\pm 1\%$ at current accuracy levels, and the true interval is likely larger due to additional run-to-run variance from random initialization and data processing. Despite these limitations in measurement, we see these results as highly encouraging for applied tasks where data may be limited and performance is critical.


\section{Discussion}
\label{sec:discussion}
In this paper we explored the space of mixed-example data augmentation, in the process generalizing and improving upon recent work~\cite{zhang2017mixup,tokozume2018image,inoue2018data}.
We sought to determine whether linearity was necessary in order for mixed-example data augmentation to be effective, and in the process of answering that question, found a surprisingly large spectrum of non-linear methods that resulted in improvements over models trained with standard augmentation methods.
Our methods, though specific to image-based tasks, are straightforward to implement and do not require any hyperparameter tuning beyond those in existing methods~\cite{zhang2017mixup,tokozume2018image}.
Though we considered a variety of methods in this work, the field of mixed-example data augmentation is still in its early stages, and we postulate that it is likely even more effective methods exist. We hope that our explorations inspire further research in the area.

Key questions for future research include developing an understanding for \emph{why} mixed-example data augmentation works and determining which specific properties of such augmentation methods are useful. We have shown that it is possible to combine the strengths of multiple approaches, but it remains unclear what the limits of mixed-example data augmentation are.
On a lower level, it would also be interesting to understand the relationship between mixed-example data augmentation and other more traditional forms of data augmentation. For example, we found the puzzling behavior that mixed-example data augmentation is only effective when performed after other forms of data augmentation, and even this simple detail eludes current understanding.

One disadvantage of our approach is that, unlike some prior work~\cite{zhang2017mixup,tokozume2018image}, our methods operate only on images. While this is true, we believe it is likely that domain-specific approaches such as ours can be made for other problems, such as speech~\cite{hinton2012deep} or natural language processing~\cite{collobert2011natural}. Furthermore, we believe that such approaches are actually most important for domain-specific tasks, such as robotics~\cite{levine2016end}, which also tend to be the most data-starved and in need of improved methodology and further research.

{\small
\bibliographystyle{ieee}
\bibliography{references}

\begin{thebibliography}{10}\itemsep=-1pt

\bibitem{abadi2016tensorflow}
M.~Abadi, P.~Barham, J.~Chen, Z.~Chen, A.~Davis, J.~Dean, M.~Devin,
  S.~Ghemawat, G.~Irving, M.~Isard, et~al.
\newblock Tensorflow: A system for large-scale machine learning.
\newblock In {\em OSDI}, volume~16, pages 265--283, 2016.

\bibitem{collobert2011natural}
R.~Collobert, J.~Weston, L.~Bottou, M.~Karlen, K.~Kavukcuoglu, and P.~Kuksa.
\newblock Natural language processing (almost) from scratch.
\newblock {\em Journal of Machine Learning Research}, 12(Aug):2493--2537, 2011.

\bibitem{devries2017improved}
T.~DeVries and G.~W. Taylor.
\newblock Improved regularization of convolutional neural networks with cutout.
\newblock {\em arXiv preprint arXiv:1708.04552}, 2017.

\bibitem{esteva2017dermatologist}
A.~Esteva, B.~Kuprel, R.~A. Novoa, J.~Ko, S.~M. Swetter, H.~M. Blau, and
  S.~Thrun.
\newblock Dermatologist-level classification of skin cancer with deep neural
  networks.
\newblock {\em Nature}, 542(7639):115, 2017.

\bibitem{gastaldi2017shake}
X.~Gastaldi.
\newblock Shake-shake regularization.
\newblock {\em arXiv preprint arXiv:1705.07485}, 2017.

\bibitem{griffin2007caltech}
G.~Griffin, A.~Holub, and P.~Perona.
\newblock Caltech-256 object category dataset.
\newblock 2007.

\bibitem{gu2017melanoma}
Y.~Gu, J.~Zhou, and B.~Qian.
\newblock Melanoma detection based on mahalanobis distance learning and
  constrained graph regularized nonnegative matrix factorization.
\newblock In {\em Applications of Computer Vision (WACV), 2017 IEEE Winter
  Conference on}, pages 797--805. IEEE, 2017.

\bibitem{gulshan2016development}
V.~Gulshan, L.~Peng, M.~Coram, M.~C. Stumpe, D.~Wu, A.~Narayanaswamy,
  S.~Venugopalan, K.~Widner, T.~Madams, J.~Cuadros, et~al.
\newblock Development and validation of a deep learning algorithm for detection
  of diabetic retinopathy in retinal fundus photographs.
\newblock {\em Jama}, 316(22):2402--2410, 2016.

\bibitem{he2016identity}
K.~He, X.~Zhang, S.~Ren, and J.~Sun.
\newblock Identity mappings in deep residual networks.
\newblock In {\em European Conference on Computer Vision}, pages 630--645.
  Springer, 2016.

\bibitem{hinton2012deep}
G.~Hinton, L.~Deng, D.~Yu, G.~E. Dahl, A.-r. Mohamed, N.~Jaitly, A.~Senior,
  V.~Vanhoucke, P.~Nguyen, T.~N. Sainath, et~al.
\newblock Deep neural networks for acoustic modeling in speech recognition: The
  shared views of four research groups.
\newblock {\em IEEE Signal Processing Magazine}, 29(6):82--97, 2012.

\bibitem{huang2016deep}
G.~Huang, Y.~Sun, Z.~Liu, D.~Sedra, and K.~Q. Weinberger.
\newblock Deep networks with stochastic depth.
\newblock In {\em European Conference on Computer Vision}, pages 646--661.
  Springer, 2016.

\bibitem{inoue2018data}
H.~Inoue.
\newblock Data augmentation by pairing samples for images classification.
\newblock {\em arXiv preprint arXiv:1801.02929}, 2018.

\bibitem{ioffe2015batch}
S.~Ioffe and C.~Szegedy.
\newblock Batch normalization: Accelerating deep network training by reducing
  internal covariate shift.
\newblock {\em arXiv preprint arXiv:1502.03167}, 2015.

\bibitem{krizhevsky2009learning}
A.~Krizhevsky and G.~Hinton.
\newblock Learning multiple layers of features from tiny images.
\newblock 2009.

\bibitem{krizhevsky2012imagenet}
A.~Krizhevsky, I.~Sutskever, and G.~E. Hinton.
\newblock Imagenet classification with deep convolutional neural networks.
\newblock In {\em Advances in neural information processing systems}, pages
  1097--1105, 2012.

\bibitem{krogh1992simple}
A.~Krogh and J.~A. Hertz.
\newblock A simple weight decay can improve generalization.
\newblock In {\em Advances in neural information processing systems}, pages
  950--957, 1992.

\bibitem{lazebnik2006beyond}
S.~Lazebnik, C.~Schmid, and J.~Ponce.
\newblock Beyond bags of features: Spatial pyramid matching for recognizing
  natural scene categories.
\newblock In {\em null}, pages 2169--2178. IEEE, 2006.

\bibitem{levine2016end}
S.~Levine, C.~Finn, T.~Darrell, and P.~Abbeel.
\newblock End-to-end training of deep visuomotor policies.
\newblock {\em The Journal of Machine Learning Research}, 17(1):1334--1373,
  2016.

\bibitem{munawar2017spatio}
A.~Munawar, P.~Vinayavekhin, and G.~De~Magistris.
\newblock Spatio-temporal anomaly detection for industrial robots through
  prediction in unsupervised feature space.
\newblock In {\em Applications of Computer Vision (WACV), 2017 IEEE Winter
  Conference on}, pages 1017--1025. IEEE, 2017.

\bibitem{russakovsky2015imagenet}
O.~Russakovsky, J.~Deng, H.~Su, J.~Krause, S.~Satheesh, S.~Ma, Z.~Huang,
  A.~Karpathy, A.~Khosla, M.~Bernstein, et~al.
\newblock Imagenet large scale visual recognition challenge.
\newblock {\em International Journal of Computer Vision}, 115(3):211--252,
  2015.

\bibitem{simonyan2014very}
K.~Simonyan and A.~Zisserman.
\newblock Very deep convolutional networks for large-scale image recognition.
\newblock {\em ICLR}, 2015.

\bibitem{srivastava2014dropout}
N.~Srivastava, G.~Hinton, A.~Krizhevsky, I.~Sutskever, and R.~Salakhutdinov.
\newblock Dropout: A simple way to prevent neural networks from overfitting.
\newblock {\em The Journal of Machine Learning Research}, 15(1):1929--1958,
  2014.

\bibitem{szegedy2016rethinking}
C.~Szegedy, V.~Vanhoucke, S.~Ioffe, J.~Shlens, and Z.~Wojna.
\newblock Rethinking the inception architecture for computer vision.
\newblock In {\em Proceedings of the IEEE conference on computer vision and
  pattern recognition}, pages 2818--2826, 2016.

\bibitem{tokozume2018image}
Y.~Tokozume, Y.~Ushiku, and T.~Harada.
\newblock Between-class learning for image classification.
\newblock In {\em Computer Vision and Pattern Recognition}, 2018.

\bibitem{tokozume2018audio}
Y.~Tokozume, Y.~Ushiku, and T.~Harada.
\newblock Learning from between-class examples for deep sound recognition.
\newblock In {\em International Conference on Learning Representations}, 2018.

\bibitem{xie2017aggregated}
S.~Xie, R.~Girshick, P.~Doll{\'a}r, Z.~Tu, and K.~He.
\newblock Aggregated residual transformations for deep neural networks.
\newblock In {\em Computer Vision and Pattern Recognition (CVPR), 2017 IEEE
  Conference on}, pages 5987--5995. IEEE, 2017.

\bibitem{zhang2017mixup}
H.~Zhang, M.~Cisse, Y.~N. Dauphin, and D.~Lopez-Paz.
\newblock mixup: Beyond empirical risk minimization.
\newblock {\em ICLR}, 2018.

\bibitem{zhong2017random}
Z.~Zhong, L.~Zheng, G.~Kang, S.~Li, and Y.~Yang.
\newblock Random erasing data augmentation.
\newblock {\em arXiv preprint arXiv:1708.04896}, 2017.

\end{thebibliography}
}

\end{document}